%% file: icra23_main.tex
\documentclass[letterpaper, 10 pt, conference]{ieeeconf}
\usepackage[T1]{fontenc}
\usepackage[utf8x]{inputenc} 
\usepackage{cite}
\usepackage{amssymb,amsfonts}
\usepackage{blindtext}
\makeatletter
\let\NAT@parse\undefined
\makeatother
\usepackage{algorithmic}
\usepackage{graphicx}
\usepackage{textcomp}
\usepackage{mathtools}
\usepackage{changes}
\usepackage{subcaption}
\usepackage{algorithm}
\usepackage{xcolor}
\usepackage{color, soul}
\usepackage{amsmath}
\usepackage{booktabs}
\usepackage{multirow}
\usepackage{xstring}
\usepackage{hhline}
\usepackage{siunitx}
\usepackage{comment}
\usepackage{physics}
\usepackage{tikz}
\usepackage{hyperref}
\usepackage{cleveref}

\crefformat{section}{\S#2#1#3} 
\crefformat{subsection}{\S#2#1#3}
\crefformat{subsubsection}{\S#2#1#3}

\floatname{algorithm}{Procedure}

\newcommand{\argmin}{\mathop{\mathrm{argmin}}}

\definecolor{rvc}{RGB}{0, 0, 0}
\definecolor{rvc2}{RGB}{0, 0, 0}
\definecolor{rvc3}{RGB}{0, 0, 0}

\sethlcolor{lightgray}

\makeatletter
\DeclareRobustCommand{\iscircle}{\mathord{\mathpalette\is@circle\relax}}
\newcommand\is@circle[2]{%
  \begingroup
  \sbox\z@{\raisebox{\depth}{$\m@th#1\bigcirc$}}%
  \sbox\tw@{$#1\square$}%
  \resizebox{!}{\ht\tw@}{\usebox{\z@}}%
  \endgroup
}
\makeatother

\IEEEoverridecommandlockouts                              

\usepackage{caption}
\newcommand{\rom}[1]{\uppercase\expandafter{\romannumeral #1\relax}}

\title{\LARGE \bf
ORORA: Outlier-Robust Radar Odometry}

\author{Hyungtae Lim$^{1}$, \textit{Student Member, IEEE}, Kawon Han$^{1}$, \textit{Member, IEEE}, Gunhee Shin$^{2}$, \\ Giseop Kim$^{3}$, Songcheol Hong$^{1}$, \textit{Member, IEEE}, and Hyun Myung$^{1*}$, \textit{Senior Member, IEEE}
\thanks{$^*$Corresponding author: Hyun Myung}
\thanks{$^{1}$Hyungtae Lim, Kawon Han, Songcheol Hong, and Hyun Myung are with the School of Electrical Engineering, KAIST (Korea Advanced Institute of Science and Technology), Daejeon, 34141, Republic of Korea. {\tt\scriptsize \{shapelim, smh1176, schong1234, hmyung\}@kaist.ac.kr} \hfill \break 
\indent $^{2}$Gunhee Shin is a research intern in Urban Robotics Lab., KAIST, Daejeon, 34141, \textcolor{rvc}{Republic of Korea}. {\tt\scriptsize gunmaplehee@naver.com} \hfill \break  
\indent $^{3}$Giseop Kim is with NAVER LABS, Seongnam, Gyeonggi-do, \textcolor{rvc}{13561, Republic of Korea}. {\tt\scriptsize giseop.kim@naverlabs.com} \hfill \break  
\indent This work was supported by the BK21 FOUR~(Republic of Korea).}
}

\begin{document}

\captionsetup[figure]{labelformat={default},labelsep=period,name={fig.}}

\maketitle
\thispagestyle{empty}
\pagestyle{empty}

\begin{abstract}

Radar sensors are emerging as solutions for perceiving surroundings and estimating ego-motion in extreme weather conditions. Unfortunately, radar measurements are noisy and suffer from mutual interference, which degrades the performance of feature extraction and matching, triggering imprecise matching pairs, which are referred to as outliers. To tackle the effect of outliers on radar odometry, a novel outlier-robust method called \textit{ORORA} is proposed, which is an abbreviation of \textit{Outlier-RObust RAdar odometry}. 
To this end, a novel decoupling-based method is proposed, which consists of graduated non-convexity~(GNC)-based rotation estimation and anisotropic component-wise translation estimation~(A-COTE). Furthermore, our method leverages the anisotropic characteristics of radar measurements, each of whose uncertainty along the azimuthal direction is somewhat larger than that along the radial direction. As verified in the public dataset, it was demonstrated that our proposed method yields robust ego-motion estimation performance compared with other state-of-the-art methods. Our code is available at {\scriptsize \href{https://github.com/url-kaist/outlier-robust-radar-odometry}{\texttt{https://github.com/url-kaist/outlier-robust-radar-odometry}}}. 


\end{abstract}

\input{section/introduction}

\input{section/methodology}

\input{section/experiments}

\input{section/results_and_discussion_v2}

\input{section/conclusion}


\bibliographystyle{IEEEtran}
\bibliography{./icra23,./IEEEabrv}

\end{document}

%% file: section/introduction.tex
\vspace{-0.1cm}
\section{Introduction} \label{sec:intro}

In recent years, the demand for localization and perception technologies on mobile \textcolor{rvc}{vehicles} has been increased. To perceive surroundings, various sensors, such as light detection and ranging~(LiDAR) and camera sensors, have been widely adopted~\cite{lim21erasor, shan2020lio, kim2022step, behley2018efficient,  behley2019semantickitti, oh2022travel, jung2020bridge, lim2022pago, qin2018vins, lee2022vivid++, lim2022uv}. However, it has been known that these sensors are severely affected by weather, illumination conditions, and interfering substances in the air, which make them difficult to be used in extreme environments~\cite{barnes2020oxford,cen2018precise, cen2019radar, burnett2022we, zhou2022towards}. 

To overcome the aforementioned drawbacks of those sensors, radar sensors are increasingly employed in autonomous vehicles and mobile robots~\cite{burnett2021we, burnett2022boreas, burnett2022we}. The radar sensors are relatively robust to atmospheric and light conditions because the radar data are measured by electromagnetic waves~\cite{han2018differential,han2021vocal,han2021detection}. In addition, with the recent advance\textcolor{rvc}{s} in low-cost and high-performance radar sensors, several novel radar-based simultaneous localization and mapping (SLAM) frameworks were proposed~\cite{checchin2010radar, holder2019real, park2021radarfactor,hong2020radarslam, hong2022radarslam}. 

\begin{figure}[t!]
    \captionsetup{font=footnotesize}
    \centering
	\begin{subfigure}[b]{0.48\textwidth}
		\includegraphics[width=1\textwidth]{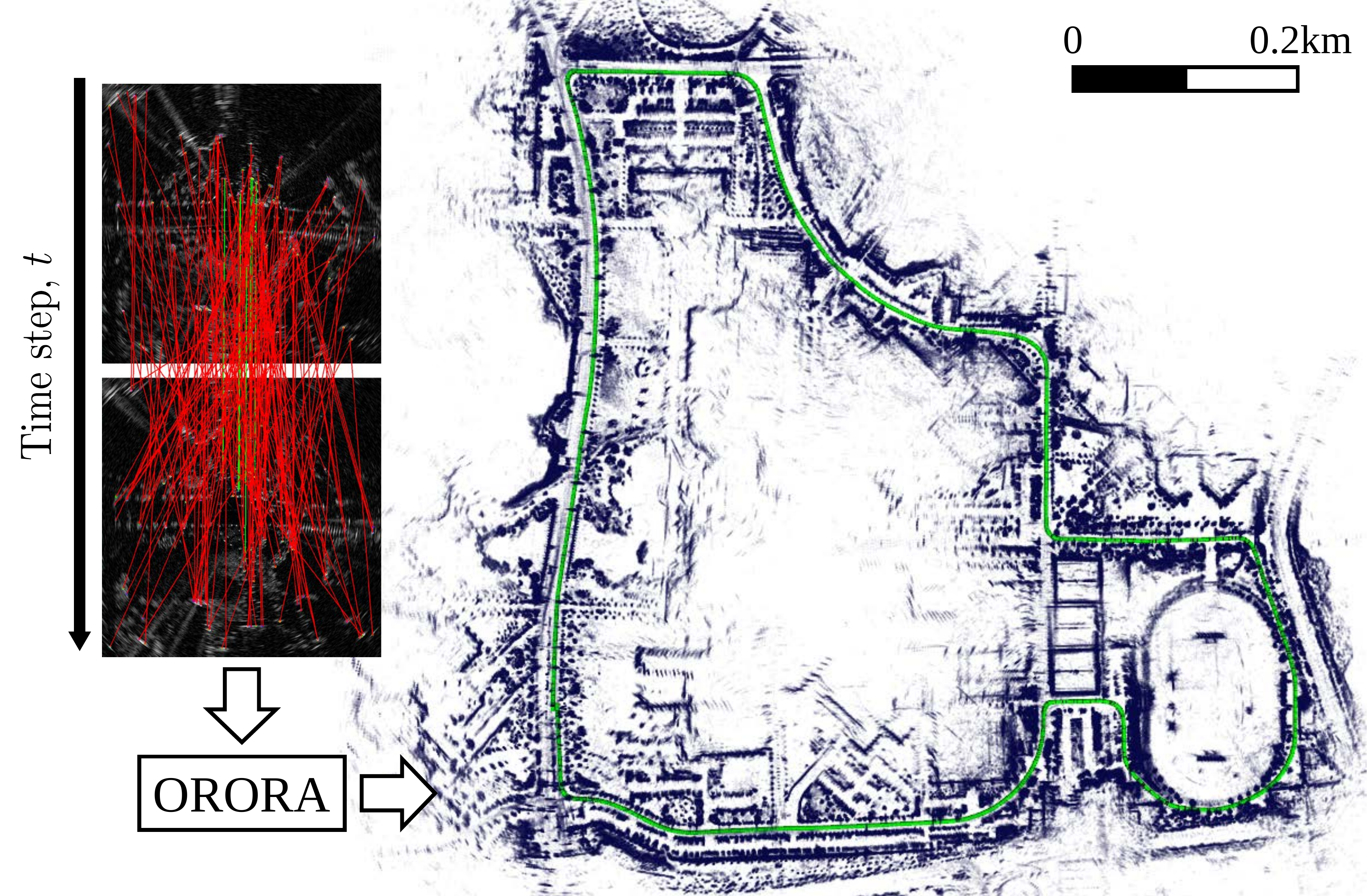}
	\end{subfigure}
	\caption{Radar odometry result of our \textit{Outlier-RObust RAdar odometry~(ORORA)} without any loop closing on a single loop of \texttt{KAIST03} in MulRan dataset~\cite{kim2020mulran}.
	Our ORORA shows robust performance even though numerous outliers are included in the estimated correspondences. The red and green lines represent the outlier and inlier feature pairs on two consecutive radar images, respectively. Note that the radar images represent an example of catastrophic failure of feature matching, which outputs only 4\% of inliers within the estimated correspondences~(best viewed in color).}
	\label{fig:fig1}
	\vspace{-0.4cm}
\end{figure}

In this context, numerous researchers have also studied radar odometry, which estimates the ego-motion of a robot when two consecutive radar data are given~\cite{park2020pharao,burnett2021radar,burnett2021we, barnes2019masking, kung2021normal, adolfsson2021cfear, aldera2019could, quist2016radar}. Radar odometry is mainly classified into two categories. One is the direct method~\cite{park2020pharao} and the other is the indirect method~\cite{burnett2021we,cen2018precise,cen2019radar, barnes2019masking}. The direct method does not use feature extraction and takes the whole radar image as an input. For instance, \textcolor{rvc}{Park \textit{et al.}~\cite{park2020pharao} proposed PhaRaO}, which is a \textcolor{rvc}{robust} direct method using phase correlation properties. Kung~\textit{et al.}~\cite{kung2021normal} modeled the uncertainty of whole radar information and exploited normal distribution transform. Barnes \textit{et al.}~\cite{barnes2019masking} proposed a deep learning-based method to achieve a novel radar matching by finding correlation in the latent space. The direct method efficiently estimates the relative poses, yet this method can be potentially affected by the integrated noises of raw radar images.

On the other hand, the indirect method exploits feature extraction and matching from the radar images. Accordingly, the indirect method takes advantage of being able to apply the techniques that are actively used in \textcolor{rvc}{the computer} vision or LiDAR fields~\cite{hong2020radarslam, hong2022radarslam, qin2018vins, cen2019radar}. Unfortunately, the indirect method also suffers from the noise of measurements and mutual interference. That is, these phenomena degrade the performance of feature extraction and matching, \textcolor{rvc}{thus} trigger\textcolor{rvc}{ing} numerous outlier pairs, as shown in Fig.~\ref{fig:fig1}.


For example, a frequency-modulated continuous-wave~(FMCW) radar sensor incorrectly detects interfering waveforms emitted by other radar signals as ghost objects. This interference also triggers speckle noise~\cite{hong2022radarslam}, which produces random noises on the radar image. Furthermore, multi-path reflection issue~\cite{han2020phase} generates one or more targets at virtual positions in the radar image. These effects also become worse once the movement of the sensor base becomes aggressive. For these reasons, the radar data are inevitably vulnerable to noises, which induces the outliers.


    
    
To tackle these noisy measurement issues, the constant false alarm rate~(CFAR) algorithm has been widely used~\cite{robey1992cfar}. As an extension, Cen and Newman~\cite{cen2018precise, cen2019radar} presented reliable feature detection methods that allow identifying objects from the radar images by using power spectra to avoid false detection. Burnett~\textit{et~al.}~\cite{burnett2021radar} and Barness~\textit{et~al.}~\cite{barnes2020under} proposed novel deep learning-based methods to extract valid feature pairs in a self-supervised manner. 
In addition, Burnett~\textit{et~al.}~\cite{burnett2021we} proposed a novel \textcolor{rvc}{random sample consensus}~(RANSAC) that compensates for \textcolor{rvc}{both} motion distortion and the Doppler effect. Hong~\textit{et~al.}~\cite{hong2020radarslam, hong2022radarslam} utilized graph-theoretic methods to prune spurious correspondences and presented feature tracking-based approach. Adolfsson~\textit{et al.}~\cite{adolfsson2021cfear} proposed a more advanced, surfel-based feature extraction method and robust point-to-line error-based optimization.

In the meanwhile, Zhou \textit{et al.}~\cite{zhou2016fast} showed that graduated non-convexity~(GNC), which gradually increases the non-linearity of the kernel as \textcolor{rvc}{the} iteration progresses to reject the effect of outliers, outperforms RANSAC~\cite{fischler1981ransac} in terms of both robustness and \textcolor{rvc}{computation} speed in LiDAR fields. Empirically, GNC robustly endures up to 70-80\% of outliers and is much faster than RANSAC~\cite{lim2022quatro}. Accordingly, GNC has been widely employed to achieve both robust and fast pose estimation~\cite{tzoumas2019outlier,yang2020teaser,sun2021iron,yang2020gnc, lim2022quatro}. In addition, Yang~\textit{et al.}~\cite{yang2020teaser} and Lim~\textit{et~al.}~\cite{lim2022quatro} demonstrated that decoupling rotation and translation estimation shows better performance when gross outliers are included in the estimated correspondences, tolerating up to 99\% of outliers.

Sharing the philosophy of GNC and the decoupling method, we propose a robust radar odometry method called \textit{ORORA}, which is an abbreviation of \textit{Outlier-RObust RAdar odometry}. To the best of our knowledge, this is the first attempt to introduce the GNC-based decoupling method in the radar fields. 

In summary, the contribution of this paper is threefold.

\begin{itemize}
	\item A novel decoupling-based outlier-robust odometry is proposed, which consists of GNC-based rotation estimation and anisotropic component-wise translation estimation~(A-COTE).
	\item To this end, the anisotropic characteristics of radar measurements, each of whose uncertainty along the azimuthal direction is larger than that along the radial direction, are mathematically modeled in the manifold space.		
	\item In experiments, our ORORA showed promising performance compared with the state-of-the-art methods \textcolor{rvc}{even though the imprecise feature} correspondences are given, which demonstrates the robustness of our ORORA against gross outliers, as presented in Fig.~\ref{fig:fig1}.  	
\end{itemize}

%% file: section/methodology.tex
\section{ORORA: Outlier-Robust Radar Odometry}

In this section, the pipeline of our radar odometry is explained, as illustrated in Fig.~\ref{fig:orora_explanation}(a). Our ORORA mainly consists of four parts: a) Doppler distortion compensation, b) graph-based outlier pruning, c) GNC-based rotation estimation, and d) A-COTE.

\subsection{Radar Image Preprocessing and Data Association}

First, we briefly explain how to estimate correspondences between two consecutive radar images. Originally, \textcolor{rvc}{the most widely used $360^\circ$ FMCW radar sensors output} a radar image whose height and width are $N_h$ and $N_w$, respectively, and each axis along the height and width denotes the radial and azimuthal direction with respect to the sensor frame, respectively. Note that the resolution in the radial direction is much better than \textcolor{rvc}{that of the azimuthal direction}, so $N_h~\gg~N_w$~(e.g., 3371 $\gg$ 400 in MulRan dataset~\cite{kim2020mulran}). 

Next, the radar image is taken as an input of the feature extraction~(in this study, the methods proposed in~\cite{cen2018precise} and \cite{cen2019radar} were used). Then, the extracted feature point on the pixel $(h,w)$ in the polar coordinates is transformed into the Cartesian coordinates as $\mathbf{p} = \color{rvc}[h c_r  \cos{\theta} ,\; h  c_r  \sin{\theta}]^\intercal$. Here, $\mathbf{p}$, $c_r$, and $\theta$ denote a 2D feature point in the Cartesian coordinates, scaling factor whose unit is meter per pixel, and radial angle, which is defined as $\theta = \frac{2\pi \cdot w}{N_w}$, respectively. Consequently, we can get the $(t-1)$-th and $t$-th feature point sets, each of which is defined as $^{t-1}\mathbf{P}$ and $^{t}\mathbf{P}$, respectively. 

\begin{figure*}[t!]
    \centering
	\begin{subfigure}[b]{0.61\textwidth}
		\includegraphics[width=1.0\textwidth]{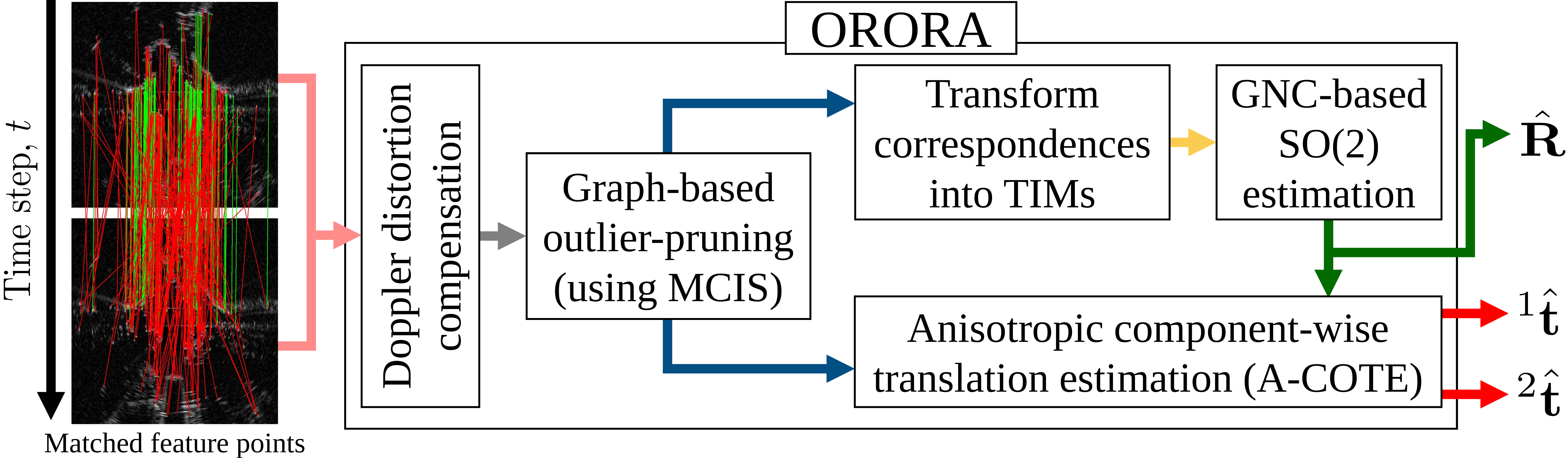}
		\caption{}
	\end{subfigure}
	\begin{subfigure}[b]{0.19\textwidth}
		\includegraphics[width=1.0\textwidth]{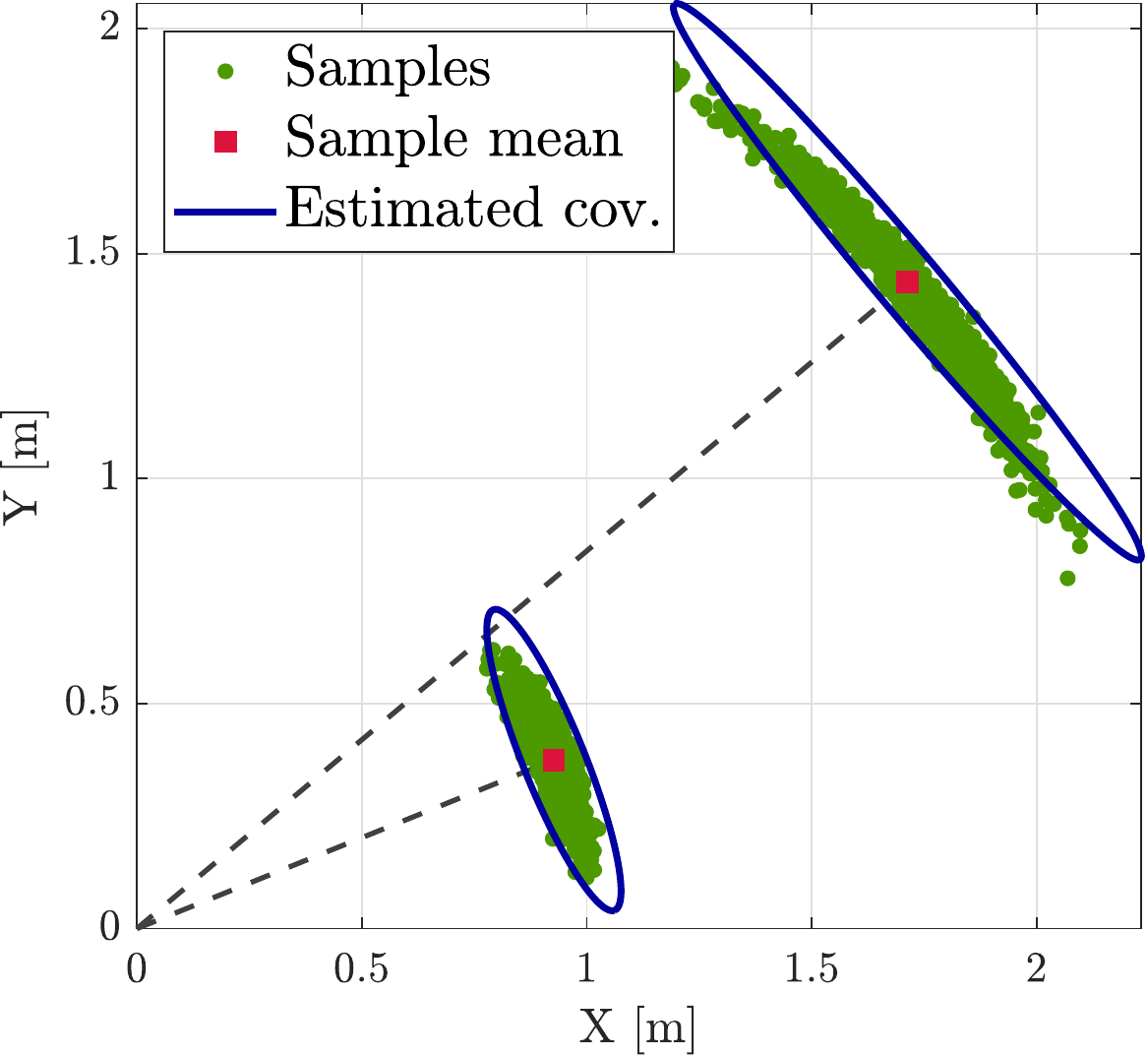}
		\caption{}
	\end{subfigure}
	\vspace{-0.2cm}
	\captionsetup{font=footnotesize}
	\caption{(a) Overall pipeline of our proposed radar odometry method. Even though the putative correspondences that include numerous outliers are given as an input~(pink arrow), our ORORA robustly estimates the relative rotation, $\hat{\mathbf{R}}$, and component-wise translation, i.e. $^1\hat{\mathbf{t}}$ for the $x$-axis and $^2\hat{\mathbf{t}}$ for the $y$-axis, respectively. (b)~Visual description of anisotropic characteristics of radar feature points. Samples (green dots) indicate that the uncertainty along the azimuthal direction is relatively larger than that along the radial direction. The blue ellipse represents the estimated covariance by our proposed mathematical modeling~(best viewed in color).}
	\label{fig:orora_explanation}
	\vspace{-0.6cm}
\end{figure*}

Next, the correspondence set between $^{t-1}\mathbf{P}$ and $^{t}\mathbf{P}$, which is denoted as $\mathcal{A}$, is estimated. To this end, ORB descriptor~\cite{rublee2011orb} is employed as proposed in~\cite{burnett2021we} to generate a descriptor of each point on the radar image transformed into the Cartesian image, as shown in Fig.~\ref{fig:orora_explanation}(a). 
Then, brute-force matching is performed, which is followed by the ratio test~\cite{lowe1999object} to initially filter out the definite outliers. 

This whole process finally outputs $\mathcal{A}$. Further details of feature extraction and matching can be found in~\cite{burnett2021we}.

\subsection{Problem Definition of Radar Odometry}

Given these putative correspondences, the goal of radar odometry is to estimate relative pose of the sensor frame between two consecutive time steps, i.e. on $t$ and $t-1$, which is expressed as $\color{rvc}\mathbf{T}^{t-1}_{t} \in \mathrm{SE}(2)$. However, using all the correspondences in $\mathcal{A}$ may cause imprecise pose estimation because the false matching\textcolor{rvc}{s} inevitably exist due to some noisy measurements, such as speckle noise~\cite{hong2022radarslam} \textcolor{rvc}{that} produces random noises due to the interference of different radar waves, multi-path reflections~\cite{hong2020radarslam}, or the effect of moving objects~\cite{han2020phase}. 

Therefore, our ultimate goal can be summarized as estimating the relative pose while minimizing the influence of the outliers as follows:

\vspace{-0.4cm}

\begin{equation}
\hat{\mathbf{R}}, \hat{\mathbf{t}} =\argmin_{\mathbf{R} \in \mathrm{SO}(2), \mathbf{t} \in \mathbb{R}^{2}}  \sum_{(i,j) \in \mathcal{A} \setminus \hat{\mathcal{O}}}\rho\Big( r(\mathbf{q}_{j}-\mathbf{R} \mathbf{p}_{i}-\mathbf{t})\Big) 
\label{eqn:final_goal}
\end{equation}

\noindent where $\mathbf{R} \in \mathrm{SO}(2)$ and $\mathbf{t}\in\mathbb{R}^{2}$ are the relative rotation and translation that correspond to $\color{rvc}\mathbf{T}^{t-1}_t$, respectively; $(i, j)$ be the index pair where $i$ and $j$ denote the indices of a point in $^{t}\mathbf{P}$ and $^{t-1}\mathbf{P}$, i.e. $\mathbf{p}_i \in {^{t}\mathbf{P}}$ and $\mathbf{q}_j \in {^{t-1}\mathbf{P}}$, respectively; $\hat{\mathcal{O}}$ denotes the estimated outlier correspondences; $\rho(\cdot)$ denotes surrogate function \cite{tzoumas2019outlier} to reduce the unintended effect caused by outliers and $r(\cdot)$ denotes the squared residual function, i.e. $||{\cdot}||^2$.

\subsection{Anisotropic Uncertainty Modeling of Radar Data}

Before we explain our proposed method, the uncertainty of each point is modeled to achieve more radar-friendly ego-motion estimation. Unlike the LiDAR measurements whose uncertainties are isotropic, which means their variances along the axes are likely to be equal to each other~\cite{yang2020teaser, lim2022quatro}, radar measurements have anisotropic characteristics. That is, the uncertainty of each point along the azimuthal direction is relatively larger than that along the radial direction; thus, the variances along the $x$-axis and $y$-axis are not equal anymore. This is due to the longer wavelength of radar signal compared with that of laser signal of a LiDAR sensor~\cite{burnett2022we}. 

To model the uncertainty of a radar feature point, the measurement noise is expressed as the superposition of a)~radial uncertainty along the direction in which the radar signal is emitted and b)~azimuthal uncertainty whose direction is perpendicular to the radial direction. Accordingly, the uncertainty is presented as a banana-like shape (green dots in  Fig.~\ref{fig:orora_explanation}(b)). 

Based on these observations, let us \textcolor{rvc2}{denote the $k$-th $\mathbf{p}$} as $\mathbf{p}_k = \gamma_k\mathbf{u}_{k} = \gamma_k [\cos{\theta_k} \, \sin{\theta_k}]^{\intercal}$ where $\gamma_k \in \mathbb{R}$, $\mathbf{u}_{k} \in \mathbb{R}^2$, and $\theta_k$ denote the measured distance, \textcolor{rvc}{a} normalized direction vector satisf\textcolor{rvc}{ying} $\left\| \mathbf{u}_{k} \right\|^2=1$, and $\theta_k = \text{atan2}(y_k, x_k)$. Here, $x_k$ and $y_k$ denote the $x$ and $y$ values of $\mathbf{p}_k$, respectively. Then, the relation between the true range value, $\gamma^{\mathrm{true}}_{k}$, and $\gamma_{k}$ is defined as follows:

\vspace{-0.25cm}
\begin{equation}
    \gamma_{k}=\gamma^{\mathrm{true}}_{k} \, + \, \delta_{\gamma_{k}}
    \label{eq:range_uncertainty}
\end{equation}
\vspace{-0.1cm}

\noindent where $\delta_{\gamma_{k}} \sim \mathcal{N}\left(0, \sigma^2_{\gamma_{k}}\right)$ is \textcolor{rvc}{a} noise term and $\sigma_{\gamma_{k}}$ denotes its standard deviation.

Second, the azimuthal uncertainty is modeled. Let $\mathbb{S}$ be the tangential space that is perpendicular to the radial direction and ${\delta}_{\mathbf{u}_{k}}$ be noise term in $\mathbb{S}$ which follows ${\delta}_{\mathbf{u}_{k}} \sim \mathcal{N}\left(0, \sigma^2_{\mathbf{u}_{k}}\right)$, where $\sigma_{\mathbf{u}_{k}}$ denotes the standard deviation in terms of the azimuthal direction. Because ${\delta}_{\mathbf{u}_{k}}$ induces unintended perturbation of rotation, the relation between the true direction vector, $\mathbf{u}^{\mathrm{true}}_{k}$, and $\mathbf{u}_{k}$ is expressed as follows:

\vspace{-0.25cm}
\begin{equation}
	\begin{aligned}
	\mathbf{u}_{k}=\mathbf{u}^{\mathrm{true}}_{k} \boxplus_{\mathbb{S}} {\delta}_{\mathbf{u}_{k}} & =  \exp({\delta}_{\mathbf{u}_{k}}\mathbf{B}) \mathbf{u}^{\mathrm{true}}_{k} \\ & \simeq (\mathbf{I}_{2} + {\delta}_{\mathbf{u}_{k}}\mathbf{B}) \mathbf{u}^{\mathrm{true}}_{k}
	\end{aligned}
	\label{eq:rot_uncertainty}
\end{equation}
\vspace{-0.15cm}

\noindent where $\boxplus_{\mathbb{S}}$ denotes the boxplus operation which is an equivalent operator of summation between 2D dimensional vector space and its manifold, i.e. $\boxplus_{\mathbb{S}}$:$~\mathbb{R}^2\times \mathbb{S} \rightarrow \mathbb{R}^2$; exp$(\cdot)$ denotes matrix exponential, which maps $\mathfrak{so}(2)$ to $\mathrm{SO}(2)$ space; $\mathbf{I}_{2}$ denotes $2\times2$ identity matrix; $\mathbf{B}=\left[\begin{array}{cc}
	0 & -1 \\
	1 & 0 \end{array}\right]$ denotes skew-symmetric matrix, which maps scalar into $\mathfrak{so}(2)$ space.

Finally, the uncertainty and covariance of each point is derived by \textcolor{rvc}{approximating} $\gamma_k\mathbf{u}_{k} \simeq (\gamma^{\mathrm{true}}_{k} \, + \, \delta_{\gamma_{k}})(\mathbf{I}_{2} + {\delta}_{\mathbf{u}_{k}}\mathbf{B})\mathbf{u}^{\mathrm{true}}_k$, based on (\ref{eq:range_uncertainty}) and (\ref{eq:rot_uncertainty}). Consequently, the uncertainty,~$\boldsymbol{\delta}_ { \mathbf {p}_k}$, and its covariance,~$\mathbf{C}_{\mathbf{p}_{k}}$, are expressed respectively as follows:

\vspace{-0.35cm}
\begin{align}
    \boldsymbol{\delta}_ { \mathbf {p} _ {k} } & =  \begin{array}{ll}
[\mathbf{u}_{k} & \left.{\gamma_{k}\mathbf{B}\mathbf{u}_{k}} \right]
\end{array}
\left[\begin{array}{l}
\delta_{\gamma_{k}} \\
{\delta}_{\mathbf{u}_{k}} 
\end{array}\right], \\
\mathbf{C}_{\mathbf{p}_{k}}  & = \mathbf{A}_{k}\left[\begin{array}{cc}
\sigma^{2}_{\gamma_k} & 0 \\
0 & \sigma^{2}_{\mathbf{u}_{k}}
\end{array}\right] \mathbf{A}_{k}^{\intercal}
\label{eq:uncertainty_each_point}
\end{align}
\vspace{-0.1cm}

\noindent where $\mathbf{A}_k \in \mathbb{R}^{2 \times 2}$ \textcolor{rvc}{represents} $\left[\mathbf{u}_{k} \;  \gamma_{k} \mathbf{B} \mathbf{u}_{k} \right]$ for simplicity. Visual description of $\mathbf{C}_{\mathbf{p}_{k}}$ is represented by the blue ellipse in Fig.~\ref{fig:orora_explanation}(b). The modeled $\mathbf{C}_{\mathbf{p}_{k}}$ is directly utilized in the initialization of rotation estimation and A-COTE~(see Sections~\rom{2}.\textit{E} and~\rom{2}.\textit{F}). 

\subsection{Overview of ORORA}

In this subsection, the procedure of ORORA is briefly explained. After the feature extraction and matching, Doppler distortion is compensated by using the estimated relative pose in the previous step, i.e. $\color{rvc}\hat{\mathbf{T}}^{t-2}_{t-1}$. Based on the constant velocity assumption, Doppler-compensated range is \textcolor{rvc}{obtained by adding} $\Delta \gamma_k^{\texttt{DPLR}}=\beta \left(v_{k, x} \cos (\theta_k)+ v_{k, y} \sin (\theta_k)\right)$ where $\beta$ denotes \textcolor{rvc}{a} constant parameter depending on the \textcolor{rvc}{modulating} frequency of a radar sensor~\cite{burnett2021we}; $v_{k, x}$ and $v_{k, y}$ denote the velocity terms of the previous movement along the $x$- and $y$-axes, respectively. \textcolor{rvc}{In our study, we only deal with autonomous vehicles which follow the non-holonomic motion,} so it can be assumed to be $v_{k, y} \simeq 0$. 
Finally, the range of each point included in $\mathcal{A}$ is corrected by adding $\Delta \gamma_k^{\texttt{DPLR}}$.

Next, max clique inlier selection~(MCIS)~\cite{rossi2013pmc} is exploited, which is a graph pruning-based outlier rejection method~\cite{hong2020radarslam, lim2022quatro}, to estimate initial outlier pairs, $\hat{\mathcal{O}}_\text{init}$. Then, these filtered correspondences, i.e. $\mathcal{A} \setminus \hat{\mathcal{O}}_\text{init}$, are taken as input\textcolor{rvc}{s for the decoupled} rotation and translation estimation, respectively~(blue arrows in Fig.~\ref{fig:orora_explanation}(a)). 

\textcolor{rvc}{Note that there are} two main differences between the previous work~\cite{hong2020radarslam}, which also uses MCIS, and ours in pose estimation procedure. First, the authors~\cite{hong2020radarslam} assumed that there are no outliers in $\mathcal{A} \setminus \hat{\mathcal{O}}_\text{init}$, so they estimate relative pose via singular value decomposition~\cite{challis1995svdpose}. In contrast, our proposed method estimates the outliers again during the rotation and translation estimation, respectively. Second, our method also considers the anisotropic characteristics of radar measurements during the pose estimation. 


\subsection{Graduated Non-Convexity-Based Rotation Estimation}

Our method is a decoupled method, so $\hat{\mathbf{R}}$ is firstly estimated, which is followed by A-COTE. To estimate $\hat{\mathbf{R}}$, two consecutive pairs ($i$, $j$) and ($i^\prime$, $j^\prime$)\textcolor{rvc2}{~=~}($i+1$, $j+1$) in $\mathcal{A} \setminus \mathcal{\hat{O}}_{\text{init}}$ are subtracted to offset \textcolor{rvc2}{the effect of translation, i.e.~$\mathbf{t}$}. By doing so, the feature pairs are expressed as translation invariant measurements~(TIMs)~\cite{yang2020teaser} as follows: $\boldsymbol{\alpha}_m~=~\mathbf{p}_{i^{\prime}} - \mathbf{p}_{i}$
for the $m$-th TIM of ${^t\mathbf{P}}$
and $\boldsymbol{\beta}_m~=~\mathbf{q}_{j^{\prime}} - \mathbf{q}_{j}$ for that of ${^{t-1}\mathbf{P}}$. Note that the effect of translation is cancelled only if both two pairs are true correspondences. Otherwise, the inherent error becomes much larger (of course, the effect of these incorrect measurements is rejected by the following procedure).

Next, GNC with a truncated least square~\cite{tzoumas2019outlier,lim2022quatro} is exploited. To this end, the objective function is first written by leveraging Black-Rangarajan duality as follows~\cite{yang2020teaser, lim2022quatro}:

\vspace{-0.45cm}
\begin{equation}
\hat{\mathbf{R}} = \argmin _{\mathbf{R} \in \mathrm{SO}(2), \\ w_{m} \in[0,1]} \sum_{m=1}^{M}\left[w_{m} r\left(\boldsymbol{\alpha}_m, \boldsymbol{\beta}_m, \mathbf{R}\right)+\Phi_{\rho_{\mu}}\left(w_{m}\right)\right]
\label{eq:black_rangarajan}
\end{equation}

\noindent where $w_m$ denotes the weight for the $m$-th TIM pair, \textcolor{rvc}{$r(\cdot)$ is the squared residual function}, and $M$ is equal to the cardinality of $\mathcal{A}\setminus\hat{\mathcal{O}}_\text{init}$. $\Phi_{\rho_{\mu}}\left(w_{m}\right)=\frac{\mu\left(1-w_{m}\right)}{\mu+w_{m}} \bar{c}^{2}$ is a penalty term \cite{yang2020gnc} where $\mu$ denotes the control parameter to increase non-convexity and $\bar{c}$ denotes \textcolor{rvc}{a} truncation parameter. 

However, solving the equation while simultaneously weighting the measurement pairs does not guarantee the optimality~\cite{zhou2016fast}. For this reason, (\ref{eq:black_rangarajan}) is iteratively solved by using alternating optimization as follows:

\vspace{-0.3cm}

\begin{equation}
\hat{\mathbf{R}}^{(n)}=\underset{\mathbf{R}\in \mathrm{SO}(2)}{\argmin} \sum_{m=1}^{M} \hat{w}^{(n-1)}_{m} r\left(\boldsymbol{\alpha}_m, \boldsymbol{\beta}_m, \mathbf{R}\right),
\label{eq:rotation}
\end{equation}

\vspace{-0.45cm}

\begin{equation}
\hat{\mathbf{W}}^{(n)}=\underset{w_{m} \in[0,1]}{\argmin } \sum_{m=1}^{M}\left[w_{m} r(\boldsymbol{\alpha}_m, \boldsymbol{\beta}_m, \hat{\mathbf{R}}^{(n)})+\Phi_{\rho_{\mu}}\left(w_{m}\right)\right]
\label{eq:weight}
\end{equation}

\noindent where the superscript~$\color{rvc}n$ denotes the $n$-th iteration and $\hat{\mathbf{W}}^{(n)}=\operatorname{diag}\left(\hat{w}^{(n)}_{1}, \hat{w}^{(n)}_{2}, \ldots, \hat{w}^{(n)}_{M}\right)$. Each $\hat{w}^{(n)}_{m}$ can be solved in a truncated closed form as follows:

\vspace{-0.35cm}

\begin{equation}
\hat{w}_{m}^{(n)}= \begin{cases}0 & \text { if } \hat{r}_{m} \in\left[\frac{\mu+1}{\mu} \bar{c}^{2},+\infty\right) \\ 
	\bar{c} \sqrt{\frac{\mu(\mu+1)}{\hat{r}_{m}}}-\mu & \text { if } \hat{r}_{m} \in\left[\frac{\mu}{\mu+1} \bar{c}^{2}, \frac{\mu+1}{\mu} \bar{c}^{2}\right) \\ 
	1 & \text {\color{rvc2} \ otherwise }\end{cases}
\label{eq:truncated_weight}
\end{equation}

\vspace{-0.15cm}

\noindent where $\hat{r}_m$ denotes $r(\boldsymbol{\alpha}_m, \boldsymbol{\beta}_m, \hat{\mathbf{R}}^{(n)})$ for simplicity. For each iteration, $\mu$ is updated as $\mu^{(n)} \leftarrow \kappa \cdot \mu^{(n-1)}$ where $\kappa > 1$ is a factor that gradually increases the magnitude of non-convexity. The iteration ends if the differential of $\color{rvc}\sum_{m=1}^{M}\hat{w}_{m}^{(n)} \hat{r}_m$ becomes sufficiently small.

The main difference between ours and the previous works~\cite{yang2020gnc,yang2020teaser,lim2022quatro} is that our method initializes the control parameter and weights, i.e. $\mu^{(0)}$ and $\hat{w}_{m}^{(0)}$, by considering the uncertainty. 
The initialization consists of four steps. First, \textcolor{rvc2}{by considering the uncertainty of the $j$ and $(j+1)$-th points}, let the approximated prior weight be $\hat{w}_m^{(\textrm{-}1)} \simeq 1/\big({\eta \color{rvc} \sqrt{\gamma^2_{j} + \gamma^2_{j+1}}}\big)$ \textcolor{rvc}{based on (\ref{eq:uncertainty_each_point}), which means that the farther, the more uncertain. Here,} $\eta$ denotes a normalization term and the superscript~$\color{rvc}{(-1)}$ indicates that the weight is on the preceding step of the initialization. 
Then, by solving (\ref{eq:rotation}) with $\hat{w}_m^{(\textrm{-}1)}$, $\hat{\mathbf{R}}^{(0)}$  is estimated. 
Third, $\mu^{(0)}$ is set as $\mu^{(0)}={\bar{c}^2}/(2 \cdot {\max(r(\boldsymbol{\alpha}_m, \boldsymbol{\beta}_m, \hat{\mathbf{R}}^{(0)}))-\bar{c}^2})$\textcolor{rvc}{~\cite{yang2020teaser}}.
Finally, $w^{(0)}_m$ is set by (\ref{eq:truncated_weight}) taking $\hat{\mathbf{R}}^{(0)}$ and $\mu^{(0)}$ as input\textcolor{rvc}{s}. 


\subsection{Anisotropic Component-Wise Translation Estimation}

Finally, the relative $x$ and $y$ translation\textcolor{rvc}{s} are estimated respectively by A-COTE. Let the discrepancy of the relative translation be $\mathbf{v}_{ij}=\mathbf{q}_{j}-{\hat{\mathbf{R}}} \mathbf{p}_{i}$, then the covariance of $\mathbf{v}_{ij}$, $\mathbf{C}_{ij}$, is equal to $\mathbf{C}_{\mathbf{q}_j} + \hat{\mathbf{R}}\mathbf{C}_{\mathbf{p}_i}\hat{\mathbf{R}}^{\intercal}$, where $\mathbf{C}_{\mathbf{q}_j}$ and $\mathbf{C}_{\mathbf{p}_i}$ are calculated by~(\ref{eq:uncertainty_each_point}), respectively. Accordingly, each uncertainty of $\mathbf{v}_{ij}$ along the $x$- and $y$-axes is equal to the $1$st and $2$nd diagonal element\textcolor{rvc}{s} of $\mathbf{C}_{ij}$, which are denoted as $^1\sigma_{ij}$ and $^2\sigma_{ij}$, respectively. Note that $^1\sigma_{ij}$ and $^2\sigma_{ij}$ have different magnitudes due to the aforementioned anisotropic characteristics of radar points. For simplicity, the property that corresponds to the $l$-th element is expressed as $^l(\cdot)$, where $l=1,2$. 

Then, the component-wise relative translation is estimated in the following four steps. First, a boundary interval set is defined as ${^{l}\mathcal{E}}$ which is $2M$-tuples. That is, ${^{l}\mathcal{E}}$ consists of $M$ lower bounds, ${^{l}\mathbf{v}}_{ij} - {^l\sigma_{ij}}$, and $M$ upper bounds, $^{l}\mathbf{v}_{ij} + {^{l}\sigma_{ij}}$. It is assumed that all the elements of $^{l}\mathcal{E}$ are sorted in ascending order. Second, consensus sets are assigned by using these noise bound values. That is, the $g$-th consensus set is defined as $^{l}\mathcal{I}_g=\{(i,j)| ({^{l}\phi_{g}}-{^{l}\mathbf{v}}_{ij})^{2} \leq {^{l}\sigma_{ij}^{2}}\}$, where ${^{l}\phi_{g}}\in\mathbb{R}$ is any value that satisfies ${^{l}\mathcal{E}}(g)<{^{l}\phi_{g}}<{^{l}\mathcal{E}}(g+1)$ for $g=1,2,\dots,(2M-1)$. 

Third, the optimal relative translation value for the $g$-th consensus set,~$^{l}{\hat{\mathbf{t}}}_g$, is estimated by the weighted average if $^{l}\mathcal{I}_g$ is non-empty as follows:

\vspace{-0.25cm}

\begin{equation}
	{^{l}{\hat{\mathbf{t}}}_{g}}=\Big(\sum_{(i,j) \in ^{l}\mathcal{I}_{g}} \frac{1}{^{l}\sigma_{ij}^{2}}\Big)^{-1}\sum_{(i,j) \in ^{l}\mathcal{I}_{g}} \frac{^{l}\mathbf{v}_{ij}}{^{l}\sigma_{ij}^{2}}.
\end{equation}

\vspace{-0.15cm}


\noindent Finally, by letting the set consisting of all the ${^l\hat{\mathbf{t}}_g}$ be $\mathbf{H}$ whose size is up to $2M-1$ and the complement of ${^l\mathcal{I}_g}$ be~${^l{\mathcal{O}}_g}$, i.e. ${^l{\mathcal{I}}_g}={\mathcal{A} \setminus (\hat{\mathcal{O}}_\text{init} \cup {^l\mathcal{O}_g}})$, the global optimum,~$^l\mathbf{t}$, is estimated by selecting $\xi \in \mathbf{H}$ and ${^l\mathcal{I}_g}$ that minimize the following objective function as follows:

\vspace{-0.2cm}
\begin{equation}
^{l}{\hat{\mathbf{t}}}, {^l\hat{\mathcal{I}}_g} =\underset{\xi \in \mathbf{H}, {^l\mathcal{I}_g}}{\argmin } \sum_{(i, j) \in {^l\mathcal{I}_g}}  \left(\frac{\xi-{^l \mathbf{t}_{ij}}}{^{l}\sigma_{ij}}\right)^2 + \sum_{(i, j) \in {^l{\mathcal{O}}_g} } {^{l}\sigma_{ij}}.
\label{eq:consensus_optim}
\end{equation}
\vspace{-0.1cm}

\noindent It is noticeable that (\ref{eq:consensus_optim}) blends the best of consensus maximization~\cite{probst2019unsupervised} and weighted average. That is, the second summand is usually much larger than the first summand, so ${^l{\mathcal{I}}_g}$ with the most cardinality is likely to be chosen as ${^l\hat{\mathcal{I}}_g}$, which behaves like consensus maximization. In the meanwhile, even though some outliers, each of whose uncertainty is sufficiently large, are unintentionally included in ${^l{\hat{\mathcal{I}}}_g}$, their larger $^l\sigma_{ij}$ naturally suppresses the effect of the outliers on the component-wise relative translation estimation.

%% file: section/experiments.tex
\section{Experiments}

\subsection{Dataset}

In our experiments, \textcolor{rvc}{most of} the sequences of MulRan dataset\footnote{\href{https://sites.google.com/view/mulran-pr/home}{\texttt{https://sites.google.com/view/mulran-pr/home}}}~\cite{kim2020mulran} were used. Note that two sequences, i.e.~$\texttt{KAIST01}$ and $\texttt{Sejong01}$, were not \textcolor{rvc}{used} because the authors~\cite{kim2020mulran} officially do not provide radar data of these two sequences in the Oxford radar image format~\cite{barnes2020oxford}.

\subsection{Error Metrics}

To quantitatively evaluate our proposed method and baseline methods, the relative translation error, $t_{\text{rel}}$, and rotation error, $r_{\text{rel}}$, are used~\cite{Zhang18rpg_eval_tool}, which are renowned criteria in odometry test. Note that these terms are normalized because each ego-motion has a different amount of movement from each other. Consequently, the units of $t_{\text{rel}}$ and $r_{\text{rel}}$ are $\%$ and $\deg / 100$\;m, respectively. 

\subsection{Implementation Details and Parameter Setting} 

First, before the feature matching, ${^t}\mathbf{P}$ is voxel-sampled with the size of $\nu$ to let these points be more evenly distributed. Then, the parameters presented in~\cite{burnett2021we} were employed for feature extraction and matching procedure.

Empirically, it was found that optimal parameters can be changed depending on the environments~(see Section~\rom{4}.\textit{A}). For this reason, we set $\nu= 0.6$\;m, $\bar{c}= 0.75$, and $\sigma_{\mathbf{u}_k}=10.8^\circ$ in widely distributed environments, whereas we set $\nu= 0.8$\;m, $\bar{c}= 1.0$, and $\sigma_{\mathbf{u}_k}=1.8^\circ$ in more partially distributed environments. Other parameters are set to be $\kappa=1.4$ and $\sigma_{\gamma_k}=0.1$\;m.

%% file: section/results_and_discussion_v2.tex
\section{Experimental Results}

In our experiments, we compared our ORORA with other feature-based state-of-the-art methods. To this end, three types of RANSAC proposed in~\cite{burnett2021we} are used: \textcolor{rvc}{the} original RANSAC~\cite{fischler1981ransac} without any compensation, \textcolor{rvc}{MC-RANSAC} whose motion distortion is compensated~\cite{burnett2021we}, and \textcolor{rvc}{MC-RANSAC + \texttt{DPLR}} that both motion and Doppler distortions are compensated\cite{burnett2021we}. For a fair comparison, we used the official open-source implementations. To closely examine the performance of ego-motion estimation depending on the quality of feature extraction, two feature extraction methods proposed in~\cite{cen2018precise} and \cite{cen2019radar} were tested, which are referred to as \texttt{Cen2018} and \texttt{Cen2019}, respectively.

\begin{figure}[t!]
	\centering 
	\begin{subfigure}[b]{0.50\textwidth}
		\includegraphics[width=1.0\textwidth]{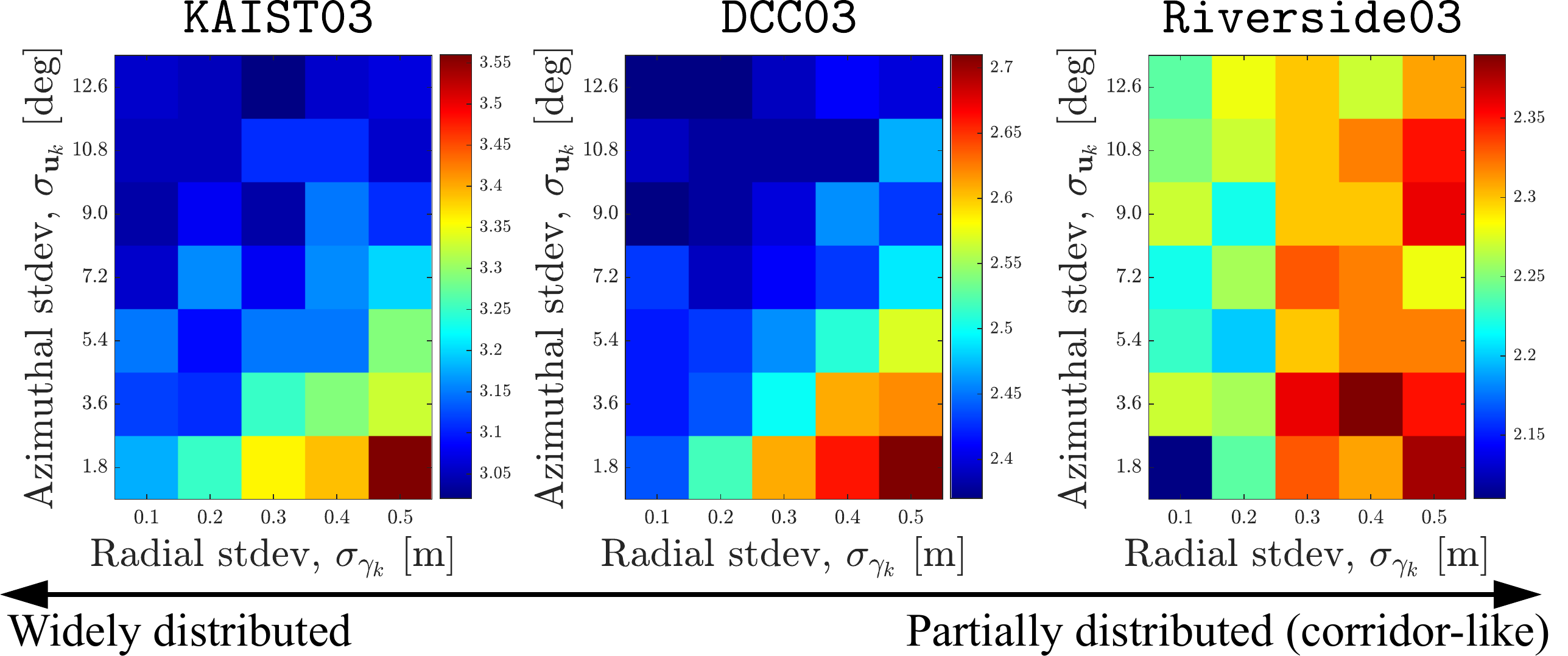}
	\end{subfigure}
	\captionsetup{font=footnotesize}
	\caption{Translation error with respect to the size of the radial and azimuthal uncertainties of each feature point. \textcolor{rvc3}{The double arrow indicates the level of distribution of feature points on the $xy$-plane}~(best viewed in color).}
	\label{fig:a-cote}
\end{figure}

\input{table/w_wo_a_cote.tex}


\subsection{Performance Changes With Respect to Uncertainty}

First, performance changes in terms of radial and azimuthal uncertainties were analyzed. Interestingly, the different tendency was observed depending on the environments, as shown in Fig.~\ref{fig:a-cote}. It was shown that larger azimuthal uncertainty leads to significant performance improvement in \textcolor{rvc3}{\texttt{KAIST03}, which includes more buildings so that feature points are widely distributed on the $xy$-plane.} The experimental results can be interpreted that the larger azimuthal uncertainty enables to suppress the potential errors caused by the feature points located \textcolor{rvc3}{distantly} on the lateral side, i.e. along the $y$-axis of the sensor frame. That is, these points are likely to have larger uncertainty along the $x$-axis due to their azimuthal uncertainties. 

In contrast, smaller azimuthal uncertainty showed better performance in \textcolor{rvc3}{\texttt{Riverside03}}. This is because the features are \textcolor{rvc3}{likely to be distributed along the $x$-axis. Accordingly, the effect of azimuthal uncertainty less affects the translation estimation}. \textcolor{rvc}{For this reason}, our A-COTE shows comparable performance with the case where anisotropic modeling is not employed, i.e. COTE~\cite{lim2022quatro}, \textcolor{rvc}{in \texttt{Riverside03}}, as shown in Table~\ref{table:w_wo_a_cote}. 


\textcolor{rvc}{Therefore, it was demonstrated that our A-COTE is more effective in complex urban environments.}

 \subsection{Comparison With the State-of-the-Art Methods}
 
 \begin{figure*}[t!]
 	\centering 
 	\begin{subfigure}[b]{0.92\textwidth}
 		\includegraphics[width=1.0\textwidth]{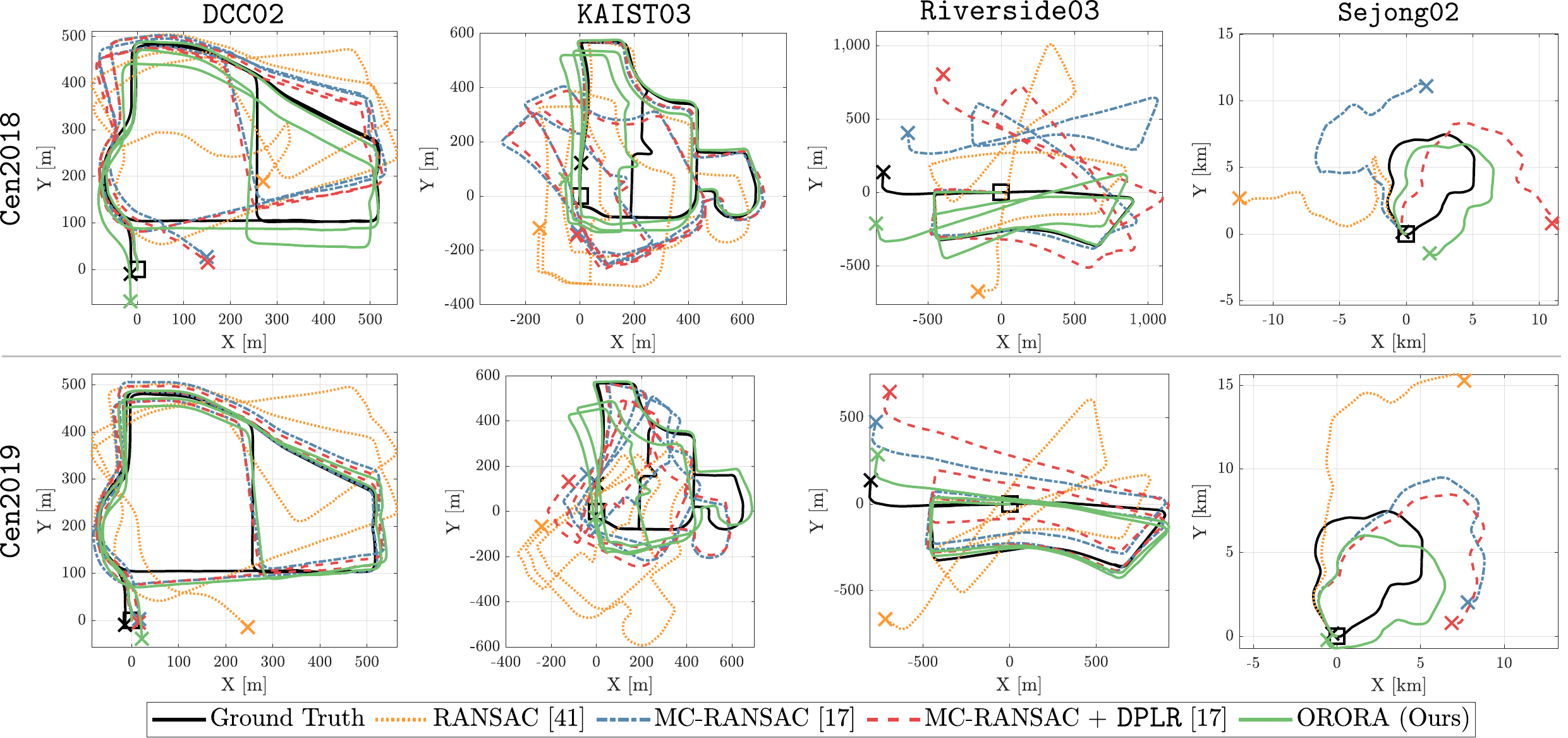}
 	\end{subfigure}
 	\vspace{-0.1cm}
 	\captionsetup{font=footnotesize}
 	\caption{Qualitative comparison with the state-of-the-art methods depending on the feature extraction methods, namely, \texttt{Cen2018}~\cite{cen2018precise} and \texttt{Cen2019}~\cite{cen2019radar}, in MulRan dataset~\cite{kim2020mulran}. $\Box$ and $\times$ marks denote start and end points, respectively~(best viewed in color).}
 	\label{fig:overall_traj}
 	\vspace{-0.2cm}
 \end{figure*}

\input{table/mulran_table_all_fine_tuning.tex}

 In general, the state-of-the-art methods showed precise odometry results, overcoming the effect of outliers. However, our ORORA showed more significant performance, as shown in Fig.~\ref{fig:overall_traj} and Table~\ref{table:mulran_comparison}. In particular, it is noticeable that no matter what feature extraction method was used, our ORORA outputs the most precise odometry performance. Therefore, it can be concluded that our ORORA is less sensitive to the performance of feature extraction methods. 
 
 Furthermore, it was observed that our ORORA is more robust against the featureless scenes. In MulRan dataset, sequences captured on the riverside and Sejong city have fewer geometrical characteristics because these scenes are wide-opened environments, so the feature extraction and matching results output less number of feature pairs and a larger ratio of outliers within the correspondences. Under that circumstance, the state-of-the-art methods showed larger performance degradation than ORORA. In addition, it was shown that once the odometry estimation becomes imprecise, the Doppler compensation occasionally rather deteriorates the odometry performance (see the performance difference of MC-RANSAC and MC-RANSAC + \texttt{DPLR}). 
 
 Therefore, these experimental results demonstrate that our ORORA is the most robust method in that our proposed method tolerates the effect of outliers and the performance of ORORA is more invariant to the quality of the estimated correspondences as well.

\subsection{Algorithm Speed}
 \vspace{-0.1cm}
 
\begin{figure}[t!]
    \centering
	\begin{subfigure}[b]{0.187\textwidth}
		\includegraphics[width=1.0\textwidth]{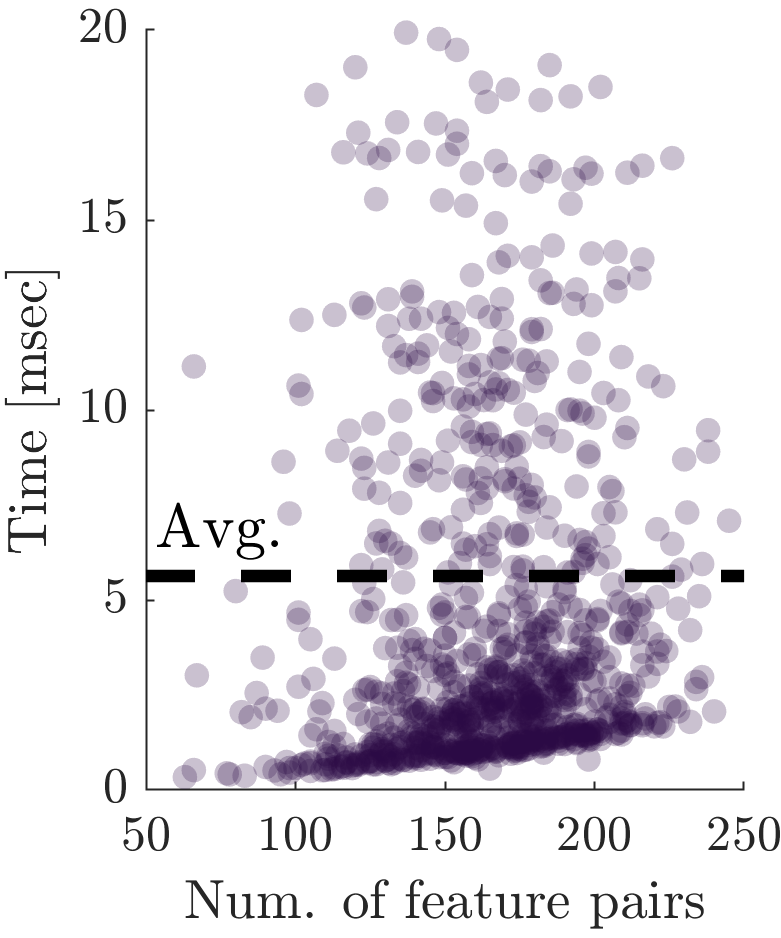}
		\caption{}
	\end{subfigure}
	\begin{subfigure}[b]{0.263\textwidth}
		\includegraphics[width=1.0\textwidth]{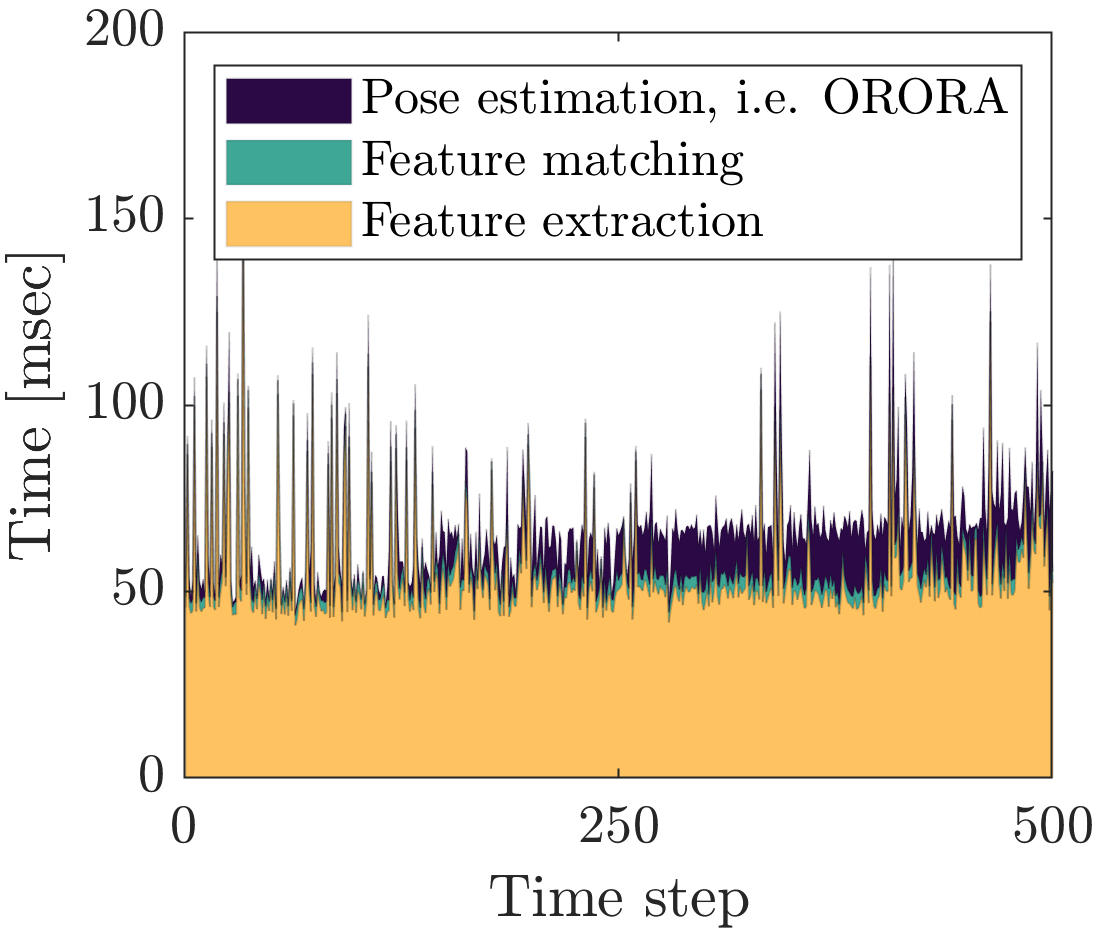}
		\caption{}
	\end{subfigure}
	\vspace{-0.15cm}
	\captionsetup{font=footnotesize}
	\caption{(a) Time taken for pose estimation with respect to the number of correspondences\textcolor{rvc}{. The average time is indicated by the black dashed line (Avg: 5.63 msec).} (b)~Area plot of computation time of each module on Intel(R) Core i7-7700K in \texttt{KAIST03} of the MulRan dataset~\cite{kim2020mulran}. As a feature extraction method, \texttt{Cen2019}~\cite{cen2019radar} was exploited~(best viewed in color).}
	\label{fig:alg_time}
    \vspace{-0.6cm}
\end{figure}

Furthermore, our ORORA is available in real-time, as presented in Fig.~\ref{fig:alg_time}. The speed of ORORA depends on the number of feature pairs, and it \textcolor{rvc}{just} took 5.63 msec on average. The whole pipeline runs at  11.5~Hz, which is sufficiently faster than the frequency at which a radar image is acquired, i.e. 4 Hz.

%% file: table/w_wo_a_cote.tex
\begingroup
\begin{table}[t!]
    \captionsetup{font=footnotesize}
	\centering
	\caption{Comparison of translation error, $t_\text{rel}$, between component-wise translation estimation~(COTE)~\cite{lim2022quatro} and \textcolor{rvc}{our A-COTE. COTE} considers the uncertainty of measurements as being isotropic. It was observed that A-COTE is more significant in obstructed environments where more buildings are located in the lateral side of the sensor frame (unit: \%).}  
	{\scriptsize
	\begin{tabular}{l|ccc}
	\toprule \midrule
	Module & \texttt{KAIST03} & \texttt{DCC03} & \texttt{Riverside03} \\ \midrule
	 COTE~\cite{lim2022quatro}     & 3.21 & 2.40 & \textbf{2.09} \\ 
	 A-COTE~(Ours)      & \textbf{3.04} & \textbf{2.37} & 2.11 \\ 
	\midrule\bottomrule
	\end{tabular}
	}
	\label{table:w_wo_a_cote}
 	\vspace{-0.3cm}
\end{table}
\endgroup

%% file: table/mulran_table_all_fine_tuning.tex
\begingroup
\begin{table*}[t!]
    \captionsetup{font=footnotesize}
	\centering
	\caption{Comparison of odometry test with the state-of-the-art methods on MulRan dataset~\cite{kim2020mulran}. Given the feature points by \texttt{Cen2018}~\cite{cen2018precise} or \texttt{Cen2019}~\cite{cen2019radar}, our ORORA outperforms the state-of-the-art methods. The bold and the gray highlight indicate the best performance among the whole method and that for each feature extraction in each sequence, respectively. All errors are represented in the form of relative translation error [\%] / relative rotation error [$\deg/100~\text{m}$].
	}
	\vspace{-0.15cm}
	\setlength{\tabcolsep}{5pt}
	{\scriptsize
	\begin{tabular}{l|l|cccccccccc}
	\toprule \midrule
	&\multirow{2}[3]{*}{Method} & \multicolumn{10}{c}{Sequence} \\  \cmidrule(lr){3-12}
	&  & \texttt{DCC01} & \texttt{DCC02} & \texttt{DCC03}  & \texttt{KAIST02} & \texttt{KAIST03} & \texttt{RIV01} & \texttt{RIV02} & \texttt{RIV03} &  \texttt{Sejong02} & \texttt{Sejong03}   \\ \midrule
 	\parbox[t]{2mm}{\multirow{4}{*}{\rotatebox[origin=c]{90}{\texttt{Cen2018}}}} & 
 	RANSAC~\cite{fischler1981ransac}         
                                        &5.11/1.10  &5.39/1.83    & 4.43/1.43 
 										&4.80/1.16& 4.94/1.42 
 										&5.19/1.27 & 7.46/1.87 & 7.31/2.21 
 										&      10.23/1.71         & 9.17/1.95  \\
 	&MC-RANSAC~\cite{burnett2021we}  
                                            &4.79/1.27&3.76/1.00& 4.39/1.40 
 											&6.60/1.74& 4.67/1.19 
 											&4.76/1.08& 6.49/1.73 & 7.72/2.19 
 											&9.18/2.14 & 8.55/1.74 \\
 	&MC-RANSAC + \texttt{DPLR}~\cite{burnett2021we}     &4.49/1.12&3.64/1.00& 4.37/1.39 
 										&6.32/1.68& 4.42/1.12 
 										&5.92/1.43&8.38/2.08& 6.61/1.88 
 										&8.83/2.32 & 12.01/2.46 \\
 	&ORORA~(Ours)              & \hl{3.18/\textbf{0.66}} & \hl{\textbf{2.38}/0.57} & \hl{2.77/0.79} 
 	                            &\hl{\textbf{3.12/0.78}} & \hl{\textbf{2.53/0.57}}
 	                            & \hl{\textbf{3.51/0.76}} & \hl{3.31/0.76} & \hl{2.79/0.64} 
 	                            & \hl{3.76/\textbf{0.67}} & \hl{5.07/0.86} \\ \midrule
 	\parbox[t]{2mm}{\multirow{4}{*}{\rotatebox[origin=c]{90}{\texttt{Cen2019}}}} &
 	RANSAC~\cite{fischler1981ransac}                  &5.41/1.39&4.16/1.07& 3.66/1.06 &4.60/1.39& 6.13/1.56 &3.99/0.96&3.54/0.98& 4.23/1.11 &4.83/1.18& 8.65/1.66  \\
 	&MC-RANSAC~\cite{burnett2021we}               &4.17/0.95&3.14/0.66& 2.70/0.67 &3.46/1.00& 3.31/0.99 &3.83/1.07&3.61/1.03& 3.94/0.96 &4.43/1.01& 6.62/1.12 \\
 	&MC-RANSAC + \texttt{DPLR}~\cite{burnett2021we}  &4.01/0.89&2.87/0.55 & 2.70/0.66 &3.59/1.08& 3.60/1.05 &4.12/1.13&3.93/1.11& 3.81/0.92 &4.51/1.01& 6.55/1.13 \\
 	&ORORA~(Ours)           &\hl{\textbf{3.12}/0.67} & \hl{2.60/\textbf{0.51}} & \hl{\textbf{2.37/0.57}} 
 	                        &\hl{3.28/0.82} & \hl{3.04/0.70} 
                            &\hl{3.53/0.84}& \hl{\textbf{2.67/0.64}} & \hl{\textbf{2.11/0.49}} 
                            &\hl{\textbf{3.27}/0.75} & \hl{\textbf{4.05/0.70}} \\ \midrule \bottomrule
	\end{tabular}
	}
	\label{table:mulran_comparison}
	\vspace{-0.4cm}
	
\end{table*}
\endgroup

%% file: section/conclusion.tex
\section{Conclusion}
 \vspace{-0.1cm}

In this study, an outlier-robust odometry method, \textit{ORORA}, has been proposed. 
The experimental results demonstrate that our ORORA successfully estimates ego-motion while rejecting the effect of outliers. In future works, we plan to study the radar-based SLAM framework by adding loop detection and loop closing modules.